\documentclass{article}

\usepackage{microtype}
\usepackage{graphicx}
\usepackage{subfigure}
\usepackage{booktabs} 
\usepackage{float}
\usepackage{amsmath}
\usepackage{amssymb}
\usepackage[subtle]{savetrees}
\usepackage{balance}

\usepackage{hyperref}

\DeclareMathOperator{\softmax}{softmax}
\DeclareMathOperator{\concat}{concat}
\DeclareMathOperator{\relu}{ReLU}
\DeclareMathOperator{\layernorm}{LayerNorm}
\DeclareMathOperator{\att}{Attend}
\DeclareMathOperator{\omni}{OmniNet}
\DeclareMathOperator{\maxpool}{MaxPool1D}
\DeclareMathOperator{\transformer}{xformer}

\DeclareMathOperator{\indexsort}{IndexSort}

\usepackage[accepted]{icml2020}

\icmltitlerunning{OmniNet: Omnidirectional Representations from Transformers}

\begin{document}

\twocolumn[
\icmltitle{OmniNet: Omnidirectional Representations from Transformers}



\icmlsetsymbol{equal}{*}

\begin{icmlauthorlist}
\icmlauthor{Yi Tay}{equal,to}
\icmlauthor{Mostafa Dehghani}{equal,to2}
\icmlauthor{Vamsi Aribandi}{to,aires}
\icmlauthor{Jai Gupta}{to}
\icmlauthor{Philip Pham}{to}
\\
\icmlauthor{Zhen Qin}{to}
\icmlauthor{Dara Bahri}{to}
\icmlauthor{Da-Cheng Juan}{to}
\icmlauthor{Donald Metzler}{to}
\end{icmlauthorlist}

\icmlaffiliation{to}{Google Research, Mountain View}
\icmlaffiliation{to2}{Google Brain Team, Amsterdam}
\icmlaffiliation{aires}{Google AI Resident}

\icmlcorrespondingauthor{Yi Tay}{yitay@google.com}
\icmlcorrespondingauthor{Mostafa Dehghani}{dehghani@google.com}

\icmlkeywords{Machine Learning, ICML}

\vskip 0.3in
]



\printAffiliationsAndNotice{\icmlEqualContribution} 

\begin{abstract}
This paper proposes Omnidirectional Representations from Transformers (\textsc{OmniNet}). In OmniNet, instead of maintaining a strictly horizontal receptive field, each token is allowed to attend to all tokens in the entire network. This process can also be interpreted as a form of extreme or intensive attention mechanism that has the receptive field of the entire width and depth of the network. To this end, the omnidirectional attention is learned via a meta-learner, which is essentially another self-attention based model. In order to mitigate the computationally expensive costs of full receptive field attention, we leverage efficient self-attention models such as kernel-based~\citep{choromanski2020rethinking}, low-rank attention~\citep{wang2020linformer} and/or Big Bird~\citep{zaheer2020big} as the meta-learner. Extensive experiments are conducted on autoregressive language modeling (LM1B, C4), Machine Translation, Long Range Arena (LRA), and Image Recognition. The experiments show that OmniNet achieves considerable improvements across these tasks, including achieving state-of-the-art performance on LM1B, WMT'14 En-De/En-Fr, and Long Range Arena. Moreover, using omnidirectional representation in Vision Transformers leads to significant improvements on image recognition tasks on both few-shot learning and fine-tuning setups.
\end{abstract}

\section{Introduction}
 Transformers~\citep{vaswani2017attention}, characterized by stacked self-attention modules and feed-forward transformations, have become a staple in modern deep learning, natural language processing~\citep{devlin2018bert,raffel2019exploring} and even computer vision~\citep{dosovitskiy2020image}. 
 One key defining characteristics in the self-attention mechanism is the global receptive field in which each token is accessible to every other token in the sequence, serving as an enabler for learning global contextual representations.
 
 This paper proposes learning omnidirectional representations from transformers. The key idea is to move beyond horizontally global receptive fields and explore the possibility of omnidirectional receptive fields. In short, we allow each token to not only attend to all other tokens in the same layer, but also all token in all the layers of the network. This global access enables tokens to have a full view of the network and as a result access the knowledge and intermediate representations of every token at each layer. 
 By modeling the relationships amongst tokens of different hierarchical levels, we are also able to capture patterns pertaining to the propagation of representations across time. Finally, this approach can be also be interpreted as a form of dense residual connection~\citep{huang2017densely}, which has been shown to be beneficial by aiding gradient flow.

Learning omnidirectional receptive fields is non-trivial for two key reasons. 
Firstly, given the quadratic complexity of the scaled dot product attention, the complexity of designing such a receptive field is increased from $N^2L$ to $(NL)^2$, where $L$ is the depth of the network and $N$ is the sequence length. We postulate that this is one big challenge that has prohibited this type of architecture from being explored in the past. 
Secondly, simply enabling omnidirectional attention from the get-go would easily cause a degeneration of the transformer into a flat network, losing much of its representational power that is enabled by sequentially refining its representations across network layers.

To mitigate these issues, our omnidirectional attention is implemented as a form of meta-learner that acts upon a standard transformer model. The meta-learner is yet another self-attention model that accepts all hidden representations across all layers as an input and refines them based on all the available information. In order to mitigate the prohibitive memory and computational costs of omnidirectional attention, we explore and evaluate multiple efficient alternatives of parameterizing the meta-learner, e.g., including fast attention via generalizable kernel attention~\citep{choromanski2020rethinking}, low-rank self-attention~\citep{wang2020linformer}, and/or block-based sparsity~\citep{zaheer2020big}.  Additionally, we further hypothesize that employing methods that try to learn the low-rank factorized structure of the entire network can lead to improved generalization capabilities - as demonstrated in our few-shot learning experiments.

Aside from varying the parameterization of the meta-learner, we also introduce partitioned variants of OmniNet in which the meta-learner is applied to blocks of $p$ consecutive layers. In short, this partitioning strategy groups the full network of $L$ layers into $\frac{L}{p}$ partitions. After computing each partition, the meta-learner learns the omnidirectional attention of all nodes across all layers in the partition.  

Via extensive experiments, we show that OmniNet achieves very promising results on a myriad of language, vision, and logic tasks. Specifically, we report strong experimental results on autoregressive language modeling \citep{chelba2013one,raffel2019exploring}, five collections of WMT machine translation, Long Range Arena~\citep{tay2020long}, and Image Recognition using Vision Transformers~\citep{dosovitskiy2020image}. Moreover, we systematically evaluate OmniNets through the lens of the performance-compute trade-off and show that they are pareto-optimal in this regard.

On machine translation, OmniNet outperforms ADMIN~\citep{liu2020very}, the current state-of-the-art 60 layers deep transformer model on two well-established machine translation collections (WMT'14 English-German and WMT'14 English-French). On the one billion language modeling benchmark, OmniNet outperforms existing state-of-the-art models such as Transformer-XL \citep{dai2019transformer}. On LRA, OmniNet improves aggregate performance over Performers \citep{choromanski2020rethinking} by $+8.9\%$ and vanilla Transformers by $+2.6\%$. On Image Recognition tasks, OmniNet demonstrates stellar few-shot learning and finetuning performance, outperforming ViT \citep{dosovitskiy2020image} by up to $\approx +3\%$ on both finetuning and few-shot learning experiments.

\section{Related Work}
Just across the past several years, attention mechanisms \citep{bahdanau2014neural} have made a significant impact on machine learning research \citep{vaswani2017attention,devlin2018bert,dosovitskiy2020image,raffel2019exploring,brown2020language,dehghani2018universal}. Simply speaking, these parameterized pooling mechanisms learn to align representations and route information based on the notion of relative importance. While early work in this area was mainly concerned with learning an alignment function between two or more sequences \citep{bahdanau2014neural,parikh2016decomposable}, there have been more focus on self-alignment (e.g., self-attention) in the recent research climate \citep{vaswani2017attention}. 

Attention mechanisms are generally applied layer-wise and operate across a one-dimensional sequence. Attention is generally bidirectional, or unidirectional in the case where a token is to be denied access to future tokens. There have been early attempts to mix information across layers in pursuit of improving gradient flow and model trainability. For example, \citep{bapna2018training} proposed transparent attention in which the decoder gains access to all encoder layers. \citep{he2018layer} proposed layer-wise coordination between encoder-decoder for machine translation. \citep{tay2018densely} proposed to densely connect the attention across stacked RNN encoder layers for language understanding tasks. The recent Realformer (residual attention) \citep{he2020realformer} proposed connecting the attention activations in a residual fashion. We believe there is sufficient evidence in the literature to suggest that mixing representations across layers is beneficial. This is further supported by fundamental work in this area such as ResNets \citep{he2016deep}, highway networks \citep{srivastava2015highway} and DenseNets \citep{huang2017densely}.

In this paper, we are mainly interested in methods for efficiently learning omnidirectional attention - an attention over the entire width and depth of the network. To this end, we leverage the recent advances in making transformers fast and efficient \citep{zaheer2020big,choromanski2020rethinking,wang2020linformer}. Many of these approaches learn an approximation via low-rank projection, kernels or block-based sparsity. An overview and extensive empirical comparison can be found at \citep{tay2020efficient,tay2020long}. To this end, the proposed approach leverages these recent advances to make what was previously not possible. By leveraging fast and efficient self-attention, we enable scalable and powerful omnidirectional attention.

\section{The Proposed Method}
This section introduces OmniNet. We first begin by reviewing the standard Transformer architecture. 
\subsection{Transformer Architectures}
This section provides a brief background for the Transformer architecture. The Transformer block accepts $N \times d$ input, where $N$ denotes the number of tokens in the sequence and $d$ denotes the size of the representation. Each Transformer block is characterized by a self-attention block and a two layered feed-forward network with ReLU activations in-between that is applied position-wise. 
\subsubsection{Self-Attention}
The self-attention mechanism first projects each input $X$ into $Q,K,V$ representations using linear transformations, corresponding to queries, keys, and values. The self-attention mechanism is typically multi-headed where multiple similar linear projections are executed in parallel. The output of each self-attention head $h$ at layer $l$ is written as:
\begin{align}
\vspace{-5pt}
y_{h,l} = \softmax \left(\frac{Q_{h,l} K_{h,l}^\top}{\sqrt{d_k}}\right) V_{h,l},   
\vspace{-5pt}
\end{align}
where $y_{h,l}$ is the output of head $h$ at layer $l$ and $d_k$ is the size of each head. The output from the multiple heads is then concatenated and then passed through another linear transformation via $W_{o,l}$ which projects the concatenation of all heads down to $d_{m}$. This is wrapped via a layer normalization followed by a residual connection and can be written as: $\layernorm(W_{o,l}\concat([y_{1,l} \cdots y_{H,l}))) + x_{l-1}$ as the final output of the self-attention module. 
\vspace{-5pt}
\paragraph{Feed Forward Layers} The FFN block of the Transformer block performs a two layer transformation defined as:
\begin{align}
\vspace{-5pt}
z_{l}=\layernorm(W_{1,l}\relu(W_{2,l}(Y))) + z_{l-1},
\vspace{-5pt}
\end{align}
where $W_1, W_2$ are trainable parameters (weight transforms) of the FFN layer. Bias parameters are omitted for clarity.
\subsection{OmniNet}

\begin{figure}
    \centering
    \includegraphics[width=1.0\linewidth]{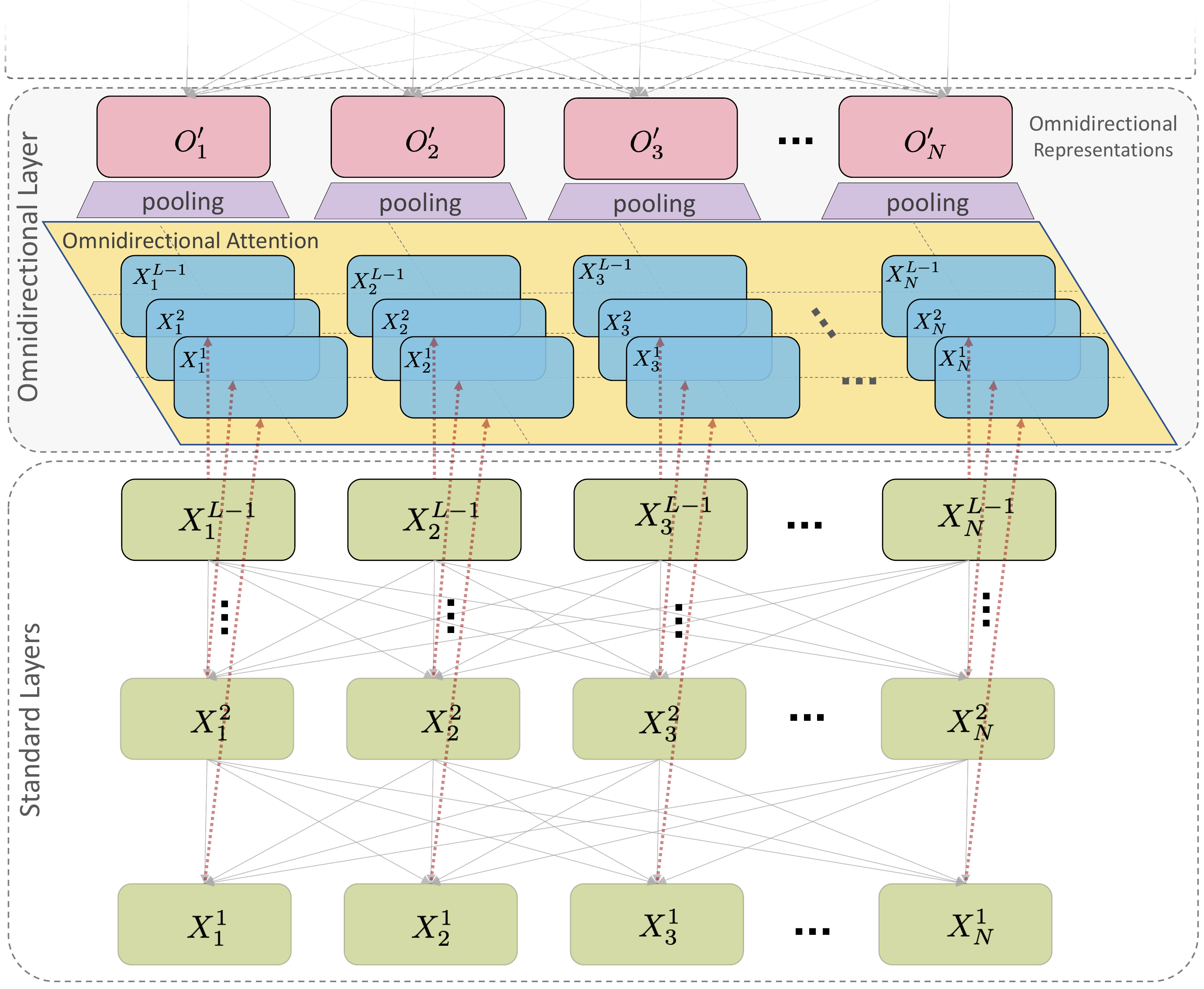}
    \vspace{-10pt}
    \caption{Overview of OmniNet. In the diagram, the omnidirectional module, when partition size is $P=L$, combines the information across all positions ($1:N$), across all layers ($1:L-1$), and for each position selects the best of all layers via a pooling operation to generate the final representations.}
    \label{fig:omninet_diagram}
    \vspace{-10pt}
\end{figure}

The proposed OmniNet method operates on an arbitrary multi-layered architecture that accepts sequential inputs. In our description, this typically refers to a stacked X-former architecture in this section. Note that while this is typically a transformer model, it can also be an arbitrary variant~\citep{choromanski2020rethinking,wang2020linformer}. Figure \ref{fig:omninet_diagram} illustrates a brief overview of the proposed OmniNet architecture.

\subsubsection{Omnidirectional Representations}
\label{sec:omni_rep}
In a stacked network of $L$ layers, each layer exposes a sequence of $N$ vectors of $d$ dimensions each. Specifically, OmniNet operates across all layers and connects the multi-layered network architecture in a \textbf{grid} like fashion. We describe the network as $\transformer$ which accepts $X$ as an input and returns a tensor of $L \times N \times d$ dimensions.
\begin{align}
\vspace{-5pt}
\transformer(X) = X_{1}, X_2 \cdots X_{L},   
\vspace{-5pt}
\end{align}
where $X_i \in \mathbb{R}^{N \times d}$. Let $X^{i}_{j}$ be the representation of $X$ at layer $i$ and position $j$ of the sequence. The OmniNet mechanism can be written as:
\begin{align}
\vspace{-5pt}
O = \att(\indexsort(X_{1}, X_2, \cdots X_{L} )),
\vspace{-5pt}
\end{align}
where $\att$ denotes an arbitrary self-attention block. The $\indexsort$ operation takes $X_1, X_2, X_L$ and sorts,\footnote{Since attention is permutation invariant this sorting simply makes it easier to (1) compute casual masks and (2) aggregate representations index-wise.} tokens within each matrix by index such that the adjacent token of the $i$\textsuperscript{th} token in layer $l$ are the $i$\textsuperscript{th} token from $l-1$ and $l+1$ respectively. 
Next, given that the input sequence length is $LN$, it is advantageous for $\att$ to be as efficient as possible. We describe three variants of OmniNet's core linear-time self-attention mechanism in subsequent sections. 

Given $O \in \mathbb{R}^{(L \times N) \times d}$, the output of the omnidirectional attention, we perform $P(.)$ a pooling operator. While there are many choices of pooling operators, parameterized or otherwise, we adopt a simple pooling function - a max pooling of stride $L$.
\begin{align}
\vspace{-5pt}
O' = \maxpool(O), 
\vspace{-5pt}
\end{align}
where $O' \in \mathbb{R}^{N \times d}$. Given $O'$, the final representation of an OmniNet augmented network is defined as:
\begin{align}
\vspace{-5pt}
\omni(X) = \transformer(X)_{L} + O'.  
\vspace{-5pt}
\end{align}
The OmniNet and main transformer model are trained together in an end-to-end fashion, i.e., gradients flow to both networks concurrently at each backward pass. 

\subsubsection{Maintaining Causality and Autoregressive Decoding}
A key point to note with $\indexsort$ is that this order enables us to apply a causal mask to the $\att$ function, namely if tokens are sorted according to sequence index first as opposed to layer first, then it would be easy to apply a causal mask $M$, where $M[i,j]=0$ when $i \leq j$ and $-\inf$ when $i>j$. This enables OmniNet to be used in autoregressive settings. 

\subsubsection{Efficient Transformers}
We describe several choices of linear-time self-attention mechanisms that are used in OmniNet's omnidirectional attention. Generally, $\att$ refers to an attention block with an attention function and a two-layered positional FFN in a similar structure to the transformer backbone. For the sake of brevity, we only describe the core attention mechanism here. Our choice of the efficient transformer is informed by \citep{tay2020long} selecting models that perform well on the compute-performance trade-off. For a list of potential variants to adopt, we refer readers to \citep{tay2020efficient}.
\vspace{-5pt}
\paragraph{Kernel-based} This variant uses the generalizable kernel attention, the fast attention mechanism proposed in Performer~\citep{choromanski2020rethinking}. Specifically, this is written as:
\begin{align*}
\vspace{-5pt}
o = W_o\concat(\hat{D_h}^{-1}(\phi(Q_h)(\phi(K_h))^\top V_h)),
\vspace{-5pt}
\end{align*}
where $\hat{D_h} = \text{diag}\phi(Q_h)((\phi(K_h))^\top 1_{L})$ and $\phi(.)$ is a random feature map that projects $\mathbb{R}^{d}$ to $\mathbb{R}^{r}$. We refer readers to~\citep{choromanski2020rethinking} for more details.
\vspace{-5pt}
\paragraph{Low-rank} Inspired by Linformer's~\citep{wang2020linformer} self-attention mechanism, we set $\att$ to be:
\begin{align*} 
\vspace{-5pt}
o = W_{o}(\concat(\softmax\left(\frac{Q_h(WK_h)^{\top}}{\sqrt{d_{k}}}\right)(WV_h)),
\vspace{-5pt}
\end{align*}
where $W \in \mathbb{R}^{N \times k}$ are low-rank projection transformations that are shared across heads and across keys and values. The complexity of this self-attention mechanism is $Nk$ instead of $N^2$, where $k \lll N$.
\vspace{-5pt}
\paragraph{Block and Memory based}
Lastly, we also explore a block and memory-based variant of efficient Transformers based on Big Bird~\citep{zaheer2020big}. In short, this is a combination of windowed attention, global attention, and sparse attention. The output for token $i$ is defined as:
\begin{align*}
\vspace{-5pt}
o_{i} = x_{i} + \sum^H_{h=1} \softmax \left( Q_{h,i}K_{h,N(i)}^\top \right) V_{h,i},   
\vspace{-5pt}
\end{align*}

where $N(i)$ is the neighborhood function which denotes the out-neighbors of node $i$, $H$ is the total number of heads and $h$ represents a head. The neighborhood function is mainly dependent on the width of the windowed attention. We refer the reader to~\citep{zaheer2020big} for more details.

\subsubsection{Partitioned OmniNet}
This section describes the types of partitioning variants that we explore in OmniNet. When $L$ is large, the eventual representation input to OmniNet can be extremely large.\footnote{A sequence length of $1K$ would result in a $11K$ input sequence length for a 12 layered Transformer model, when using an omnidirectional layer as the final layer.}

Let $P$ be an integer valued hyperparameter that determines the partition size. For a $L$ layer transformer network, when $\ell \mod ~P$ is $0$, we insert a meta-learner block. 
\[
\vspace{-5pt}
X_{\ell} = 
\begin{cases}
\att(X_{\ell - P}, \cdots X_{\ell-1})), \:\:\:\:\: \text{if } \:\:\:\:\:\: \ell \mod P = 0 \\ 
\transformer(X_{\ell - 1})
\end{cases}
\]
In short, whenever $\ell \mod ~P=0$, we activate an omnidirectional attention layer, aggregating representations all the way from the previous partition $\ell - P$ layer up till $\ell - 1$. In this case, we skip the original $\transformer$ layer, hence \textbf{maintaining approximately the same parameter size} of the network.
\section{Experiments}
\begin{table}[t]
    \vspace{-5pt}
    \centering
    \caption{Experimental results (quality, i.e., perplexity scores at 30K and 100K respectively) on autoregressive language modeling. All models are approximately 50M parameters.}
    \begin{tabular}{l|cc}
    \toprule 
      Model   & LM1B & C4 
      \\
      \midrule
      Transformer   & 33.14  & 34.86 \\
      Realformer &32.95 & 35.63 \\
      Performer & 34.33 & 35.68\\ 
      BigBird & 32.90 &  38.36\\ 
      \midrule
      OmniNet$_B$ &33.69 (-1.7\%)  & 34.73 (+0.4\%) \\
      OmniNet$_P$ & 30.19 (+9.0\%) & 33.97 (+2.6\%) \\
      OmniNet$_T$ &\textbf{30.12} (+9.1\%) & \textbf{33.39} (+4.2\%)\\
      \bottomrule
    \end{tabular}
    \label{tab:lmexp}
    \vspace{-18pt}
\end{table}

\begin{table}[t]
    \centering
    \caption{Comparison with existing state-of-the-art and published works on One Billion Word Language modeling~\citep{chelba2013one} benchmark.}
    \begin{tabular}{l|cc}
    \toprule
        Model & \#Params &PPL  \\
        \midrule
        Adaptive Input (\citeauthor{baevski2018adaptive}) & 0.5B&24.1 \\
        Adaptive Input (\citeauthor{baevski2018adaptive}) & 1.0B &23.7 \\ 
         Transformer-XL (\citeauthor{dai2019transformer}) &0.5B &23.5\\
         Transformer-XL (\citeauthor{dai2019transformer}) & 0.8B &21.8\\ 
         \midrule
         OmniNet$_P$ (Large) & 0.1B & 21.6 \\
         OmniNet$_B$ (Large) & 0.1B & 22.0\\ 
         OmniNet$_T$ (Large)  & 0.1B & \textbf{21.5}\\ 
         \bottomrule
    \end{tabular}
    \label{tab:lmsota}
    \vspace{-15pt}
\end{table}
We conduct experiments on autoregressive language modeling, machine translation, long range sequence modeling and a series of image recognition tasks. Our implementation uses Flax \citep{flax2020github} and Jax \citep{jax2018github}.


\subsection{Autoregressive Language Modeling}
We run experiments on large-scale autoregressive (unidirectional) language modeling. We use two \textbf{large-scale} datasets, language modeling one billion (LM1B)~\citep{chelba2013one} and the Colossal Cleaned Common Crawl corpus (C4)~\citep{raffel2019exploring}. 

\begin{table*}[t]
    \vspace{-12pt}
    \centering
    \caption{Results on five collections from the WMT'17 machine translation task.}
    \begin{tabular}{l|ccccc}
    \toprule
    Model & En-De & En-Fi & Cs-En & En-Fr & Ru-En  \\
    \midrule
      Transformer.   & 28.6 &  20.5 & 22.2 & 43.0 & 35.8 \\
      \midrule
         OmniNet$_L$  & 28.8 (+0.7\%)& 20.8 (+1.5\%) &  22.8 (+2.7\%)& \textbf{43.3} (+0.7\%) & 36.2 (+1.1\%) \\ 
         OmniNet$_B$  & 28.8 (+0.7\%) &20.9 (+2.0\%)  & 22.6 (+1.8\%) & 43.2 (+0.5\%)& 34.2 (-4.5\%)\\ 
           OmniNet$_P$ &\textbf{29.0} (+1.4\%) & 20.9 (+2.0\%) & \textbf{23.0} (+3.6\%) & 43.1 (+0.2\%) &36.2 (+1.1\%) \\
         \bottomrule
    \end{tabular}
    \label{tab:wmt17}
    \vspace{-17pt}
\end{table*}

\begin{table}[t]
    \vspace{-5pt}
    \centering
    \caption{Comparisons with the state-of-the-art on WMT'14 En-De  and WMT'14 En-Fr. OmniNet outperforms ADMIN~\citep{liu2020very}, the current state-of-the-art deep transformer model for MT.}
    \begin{tabular}{l|cc}
    \toprule
    Model & En-De & En-Fr \\
    \midrule
   Evolved Trans. \citep{so2019evolved}   & 29.2 & n/a \\
      Large Trans. \citep{ott-etal-2018-scaling}  & 28.6 & 41.4  \\
      60L Trans. \citep{liu2020very} & 29.5 & 41.8\\ 
      \midrule
      OmniNet$_P$ & \textbf{29.8} & \textbf{42.6}\\ 
         \bottomrule
    \end{tabular}
    \label{tab:wmt14}
    \vspace{-22pt}
\end{table}

\subsubsection{Experimental Setup} 
For both tasks, we use a max length of $256$ subword tokens per example and report scores on subword perplexity on the validation set. In the first ablative experiment, we train all models for $30K$ for LM1b and $100K$ steps for C4 using 16 TPU-V3 Chips. Models are of \textit{base} size and have an embedding dimension of $512$, $8$ heads, $6$ layers and hidden dimensions (MLP) of $2048$. For strong baselines, we compare with Transformers, Performers~\citep{choromanski2020rethinking}, and BigBird~\citep{zaheer2020big}. We also add the recent Realformer (residual attention Transformer)~\citep{he2020realformer} as a strong baseline. For OmniNet, we tune the partition amongst $\{2,3,6\}$. All models have approximately $50M$ parameters. Next, we are interested in (1) how OmniNet scales to large sizes and (2) comparing with other published works~\citep{dai2019transformer}. Hence, we implement a larger OmniNet with MLPs of size 8K and head size of 2K. 
\subsubsection{Results on Language Modeling} 
Table \ref{tab:lmexp} reports results on LM.  We observe that OmniNet$_{P,T}$ outperforms all baselines by about $+9.1\%$ on LM1b and $+4.2\%$ on C4. We also outperform strong baselines such as Realformer, BigBird, and vanilla Transformers on both corpora. We also observe that OmniNet$_P$ performs reasonably close to OmniNet$_T$, which ascertains that using an efficient Transformer may be sufficient for omnidirectional attention. On the other hand, Table \ref{tab:lmsota} reports a comparison with other published works on LM1B. Notably, OmniNet$_{P,T}$ (large) outperforms Transformer-XL and achieves state-of-the-art performance.

\subsection{Neural Machine Translation}
We conduct experiments on machine translation, a sequence-to-sequence task. for evaluating Transformer models. 
\subsubsection{Experimental Setup} 
We use five collections/datasets from WMT-17,\footnote{\url{http://www.statmt.org/wmt17/translation-task.html}} namely En-De (English $\rightarrow$ German), En-Cs (English $\rightarrow$ Czech), En-Fi (English $\rightarrow$ Finnish), En-Fr (English $\rightarrow$ French) and En-Ru (English $\rightarrow$ Russian). We train all models using $16$ TPU-V3 chips with a batch size of $256$. Our base Transformer model has 6 layers, a hidden size of $4096$, embedding size of $1024$, and a head size of $1024$. The number of heads is $16$. We use a SentencePiece \citep{kudo2018sentencepiece} vocabulary of $32K$ built for each language specifically. More details can be found in the appendix.

\subsubsection{Results on WMT'17 Machine Translation}
Table \ref{tab:wmt17} reports results on all 5 collections of WMT'17. Overall, OmniNet$_P$ outperforms the vanilla Transformer model on all five collections, with up to $\approx+3.6\%$ performance improvement. Similar to LM, we find that the performer variant works the best and the BigBird variant works the worse.
\subsubsection{Comparisons against state-of-the-art} We train a large OmniNet model and compare it with the state-of-the-art approaches. We compare with ADMIN~\citep{liu2020very}, a very deep (60 layers) Transformer model that achieves state-of-the-art performance on the WMT En-De dataset. We use a $8$ layer OmniNet model with $4096$ MLP dimensions, $1024$ hidden dimensions and embedding dimensions. We compare models using \textit{sacrebleu}~\citep{post2018call}. For OmniNet, given the strong performance of the Performer variant on WMT'17 collections, we only train a single $P$ variant OmniNet for comparing with SOTA models. Further details of the setup can be found in the appendix.

\begin{table}[t!]
    \vspace{-5pt}
    \centering
    \caption{Results on Long Range Arena~\citep{tay2020long}.}
    \begin{tabular}{c|ccc|c}
    \toprule
       Model  & Text & Retrieval & ListOps &  Avg \\
       \midrule
       Linformer & 53.9 & 52.3 & 35.7 & 47.3 \\
       BigBird &  64.0 & 54.7 & 36.1 & 51.6 \\
       \midrule
       Performer & 65.4& 53.8 & 18.0 &  45.7  \\
       +OmniNet$_P$ & \textbf{65.6} & 60.9 &  18.2 & 48.2\\
       +OmniNet$_L$ & 63.1 & \textbf{63.7} & \textbf{37.1} & 54.6 \\ 
       \midrule
       Transformer  & 62.5 & 57.5 & 36.4 & 52.1 \\ 
       +OmniNet$_P$ & \textbf{65.1} & 58.8 & \textbf{37.2} & 53.7 \\
       +OmniNet$_L$ & 63.1 & \textbf{63.8} &\textbf{37.2} & \textbf{54.7} \\ 
       \bottomrule
    \end{tabular}
    \label{tab:lra_results}
    \vspace{-22pt}
\end{table}
Table \ref{tab:wmt14} reports results on WMT'14 En-De and En-Fr. Our results show that OmniNet outperforms the existing state-of-the-art ADMIN model~\citep{liu2020very}, a 60-layer deep transformer model. 
\subsection{Long Range Arena}
We conduct experiments on the recently proposed Long Range Arena benchmark~\citep{tay2020long}. The goal of this experiment is to show that OmniNet improves long-range sequence modeling. A dual goal is  to show that it is possible to combine different inductive biases to obtain a better efficient Transformer model that is versatile on different types of data.

\begin{table*}
\vspace{-10pt}
\centering
\caption{Transfer performance of pre-trained OmniNet and equivalent ViT models in fine-tuning setup on popular image classification benchmarks. All models are pre-trained on the JFT-300M dataset and fine-tuned on the target dataset.}
\begin{tabular}{l c @{}p{0.2cm}@{} c | c @{}p{0.2cm}@{} c}
\toprule
   & ViT$_{B/32}$ &  & OmniNet$_{B/32}$ & ViT$_{B/16}$ &  & OmniNet$_{B/16}$ \\ 
\midrule 
ImageNet &  0.8073 & 	$\rightarrow$  & 0.8374 &  
            0.8415 & 	$\rightarrow$  & 0.8626 \\
CIFAR-10 &  0.9861 & 	$\rightarrow$  & 0.9900 &  
            0.9900 & 	$\rightarrow$  & 0.9940   \\
CIFAR-100 &  0.9049 & 	$\rightarrow$  & 0.9153 &  
            0.9186 & 	$\rightarrow$  & 0.9224  \\
Oxford-IIIT Pets &  0.9340 & 	$\rightarrow$  &0.9441  &  
                    0.9580 & 	$\rightarrow$  &0.9674  \\
Oxford Flowers-102 & 0.9927 & 	$\rightarrow$  &0.9954 & 
                     0.9956  & 	$\rightarrow$  &0.9961 \\
\midrule 
exaFLOPs  & 165 && 193 &  743 && 891 \\
\bottomrule
\end{tabular}
\label{tbl:OmniNet-ViT-fineting}
\end{table*}

\subsubsection{Experimental Setup}
We run two key experiments using Transformer and Performer as the main backbone model and vary the meta-learner in OmniNet, using Linformer and Performer variants of the OmniNet meta-learner. The goal is to demonstrate that OmniNet translates to backbone agnostic improvements. We run OmniNet experiments using the LRA codebase and run OmniNet models using the same hyperparameters as the results reported in~\citep{tay2020long}. Note that LRA is comprised of five benchmarks, however, we omit Image and Pathfinder experiments since the best hyperparameters on these tasks turn out to be shallow (1-2 layered) models. We report the average of the text, retrieval, and ListOps tasks.

\subsubsection{Results on LRA}
Table \ref{tab:lra_results} reports the results on our LRA experiments. Firstly, we observe that OmniNet makes substantial improvements to the base model, regardless of whether it is a Transformer or Performer. Notably, with OmniNet$_L$, the Linformer meta-learner, the Performer model is improved by almost $6$ to $7$ absolute percentage points. An interesting observation can be made on the ListOps task where Omninet$_P$ (Performer variant) does not result in much improvement over the base Performer. However, the performance doubles with OmniNet$_L$. Since the base Linformer model does pretty well on this task, we postulate that this is due to OmniNet$_L$ providing a Linformer-like inductive bias to the Performer model. Finally, we observe that OmniNet improves the vanilla Transformer in both cases ($P$ or $L$),  improving the average score by about $+2.6\%$ absolute percentage points.
\nocite{langley00}

\subsection{Image Recognition}

\begin{figure*}[t!]
\vspace{-10pt}
\begin{center}
\includegraphics[width=0.98\textwidth]{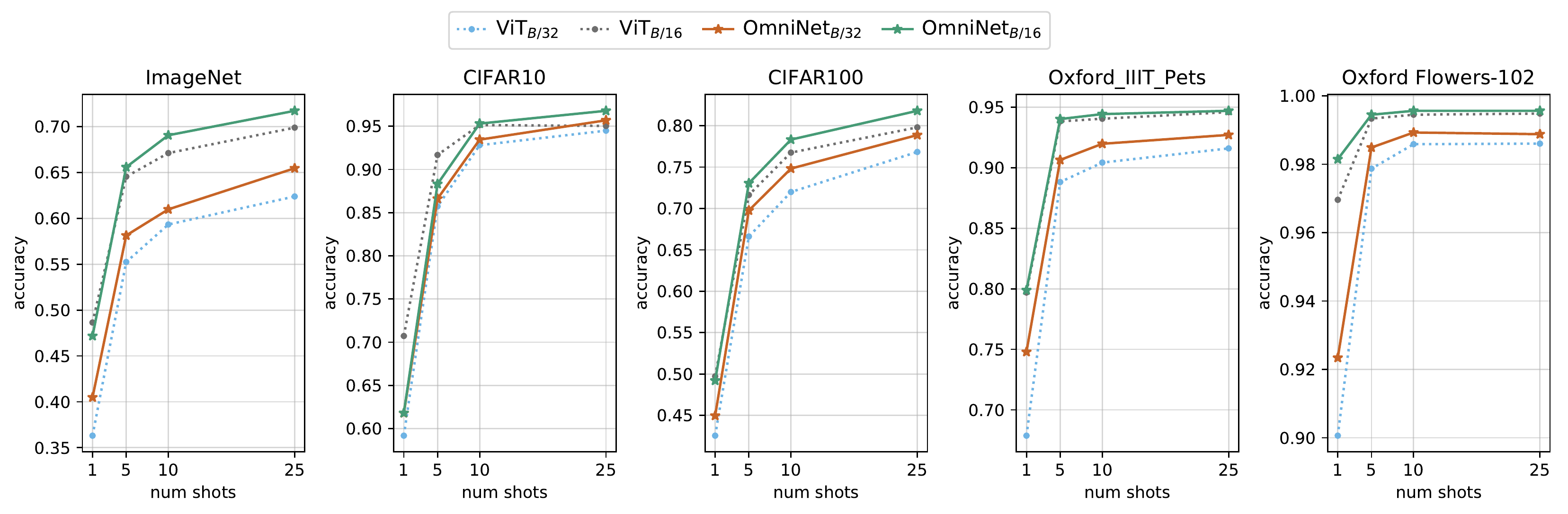}
\end{center}
\vspace{-15pt}
\caption{Performance of pre-trained OmniNet and equivalent ViT models in few-shot learning setup on downstream tasks, when transferred using only few images (1, 5, 10, and  25) per class.}
\label{fig:OmniNet-ViT-few_shot}
\vspace{-15pt}
\end{figure*}

Transformer-based models started showing competitive performance on different vision tasks like classification, object detection, and segmentation~\citep{chen2020generative, dosovitskiy2020image, carion2020end, kumar2021colorization}.

To showcase the power of omnidirectional representations in yet another task, we incorporate the omnidirectional representation in Vision Transformer (ViT)~\citep{dosovitskiy2020image}, when pre-trained on a large amount of data in a supervised fashion and evaluated on downstream image recognition tasks, either through few-shot learning or fine-tuning.

\subsubsection{Vision Transformer}
Vision Transformers (ViT)~\citep{dosovitskiy2020image} have recently shown impressive results on image classification compared to state-of-the-art convolutional networks, while they require significantly fewer computational resources to train. ViT is a standard Transformer that is directly applied to images. To do so, we first split the input images into non-overlapping patches and embedded them using a linear projection. The patch embeddings are provided as a sequence of tokens to a Transformer. 
When pre-trained on large datasets (14M-300M images) at a sufficient scale, ViT shows excellent results that are transferable to tasks with fewer data points.

\subsubsection{Experimental Setup}
Similar to the ViT setup, we pre-train our OmniNet models on the JFT dataset~\citep{sun2017revisiting} with 18k classes and 303M images, for $7$ epochs. We evaluate our models in the transfer setup (few-shot and fine-tuning) on several downstream tasks: ImageNet, CIFAR-10, CIFAR-100~\citep{krizhevsky2009learning}, Oxford-IIIT Pets~\citep{parkhi2012cats}, and Oxford Flowers-102~\citep{nilsback2008automated}. We follow the pre-processing from~\citep{kolesnikov2019big} on both upstream and downstream datasets, which is used in the original ViT experiments.

In our experiments, we train and evaluate OmniNet$_{B/32}$ and OmniNet$_{B/16}$, which are based on ViT$_{B/32}$ and ViT$_{B/16}$.\footnote{Note that SOTA results on the downstream tasks we use here are from ViT$_{H/14}$~\citep{dosovitskiy2020image}, which has more than seven times as many parameters than the \textit{base} models we use as baselines. Here, we aim at merely showcasing the gain of using omnidirectional representations in the image recognition task.} 
Similar to ViT$_{B/32}$ and ViT$_{B/16}$, OmniNet$_{B/32}$ and OmniNet$_{B/16}$ are both ``\textit{base}'' models, adopted from BERT, and use patch sizes of $32\times32$ and $16\times16$ respectively. 

In our OmniNet models, we replace the final layer of ViT with an omnidirectional layer. In other words, we set the portion size $P=12$. In this task, we limit our experiments to using Performers~\citep{choromanski2020rethinking} in the omnidirectional attention, given their strong results among the efficient transformer variants.

During pre-training, we use a batch size of $4096$ using Adam with $\beta_1=0.9$ and $\beta_2=0.999$, and use a weight decay of $0.05$ for OmniNet. We use a learning rate of $8\mathrm{e}{-4}$ with a linear decay and a linear warmup of $10K$ steps.

\subsubsection{Results on Image Recognition}
We first present the results of OmniNet and corresponding ViT models as baselines in the fine-tuning setup. For fine-tuning, we use SGD with momentum and a batch size $512$ in all downstream tasks. 
Table \ref{tbl:OmniNet-ViT-fineting} presents the results of fine-tuning experiments. We also report the total pre-training compute, in terms of number of FLOPs for each model. As we can see, introducing a module that learns omnidirectional representations to Vision Transformers leads to improvements on different downstream tasks. Given these improvements and comparing the number of FLOPs for OmniNets and ViTs, we can see that the additional compute, thanks to efficient attention techniques, is fairly reasonable.

For evaluating OmniNet in the few-shot learning setup, similar to ViT, we train a linear classifier on top of the representations from the frozen pre-trained model, given only a subset of training examples. Plots in Figure~\ref{fig:OmniNet-ViT-few_shot} illustrate the accuracy of OmniNet and ViT, using different numbers of shots. The results indicate that adding omnidirectional representations to ViT leads to better transfer across all downstream datasets.

\subsection{Effect of Partition Size and Compute/Performance Trade-offs}
OmniNet offers the flexibility to apply the Omnidirectional layers on different partition sizes. With smaller partition sizes, we attend to tokens from fewer layers, and with bigger partition, we widen the vertical receptive field of the omnidirectional attention, which might be effective for learning better representations by capturing information from more levels. In terms of computational costs, however, there is a trade-off when choosing the partition size. Small partition sizes means running attention on smaller sequences while repeating it more frequent, and bigger partition sizes means dealing with longer sequences, but having fewer omnidirectional layers in the network. 

\begin{figure}[t!]
\vspace{-11pt}
\begin{center}
\includegraphics[width=0.98\columnwidth]{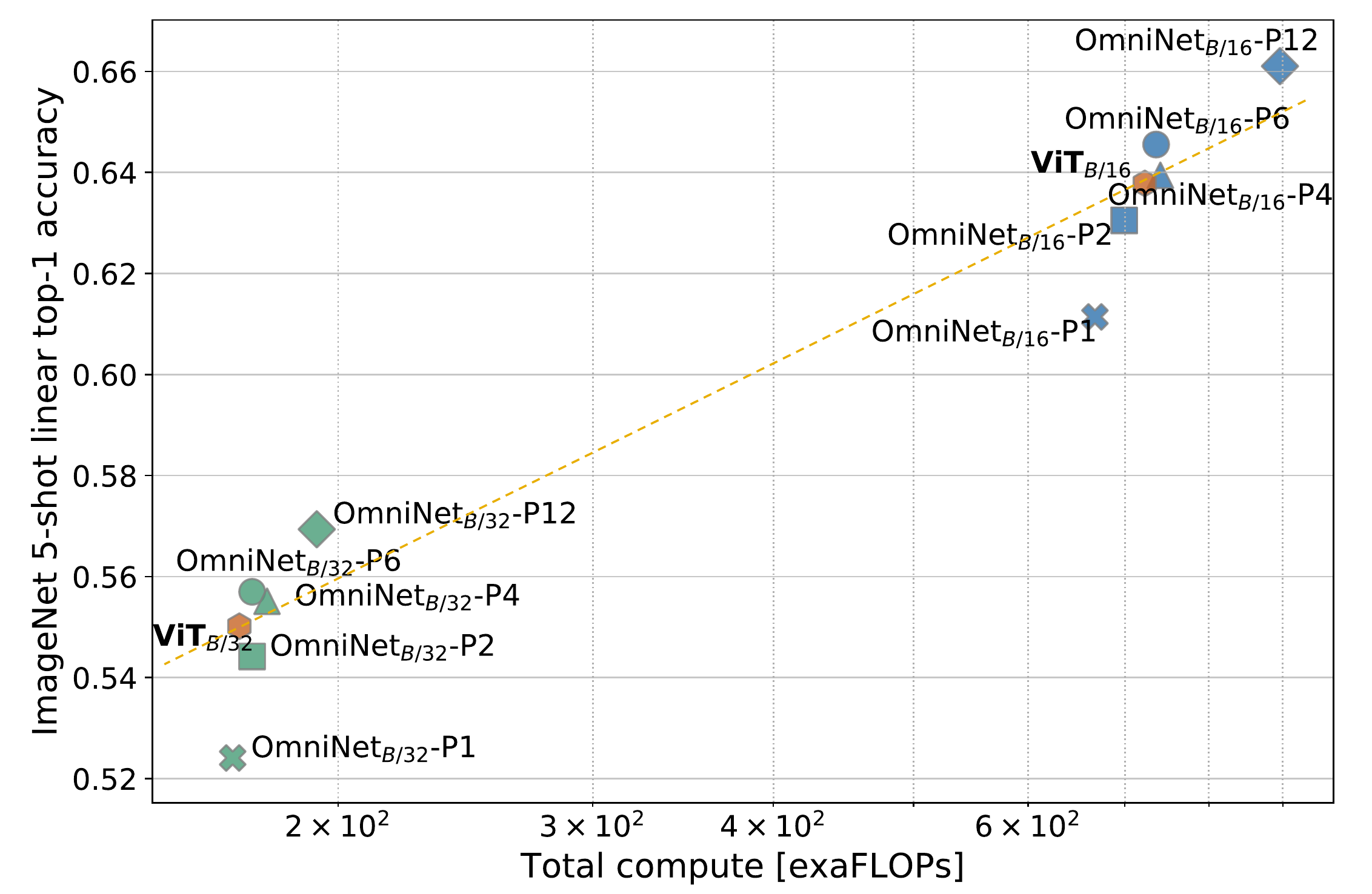}
\end{center}
\vspace{-15pt}
\caption{Performance of ViT and OmniNet (with different partition sizes) in terms of top-1 accuracy on ImageNet 5-shot linear, versus their computational costs in terms of number of FLOPs.}
\label{fig:OmniNet-ViT-ablation}
\vspace{-21pt}
\end{figure}

\begin{figure*}
    \centering
    \vspace{-5pt}
    \includegraphics[width=\linewidth]{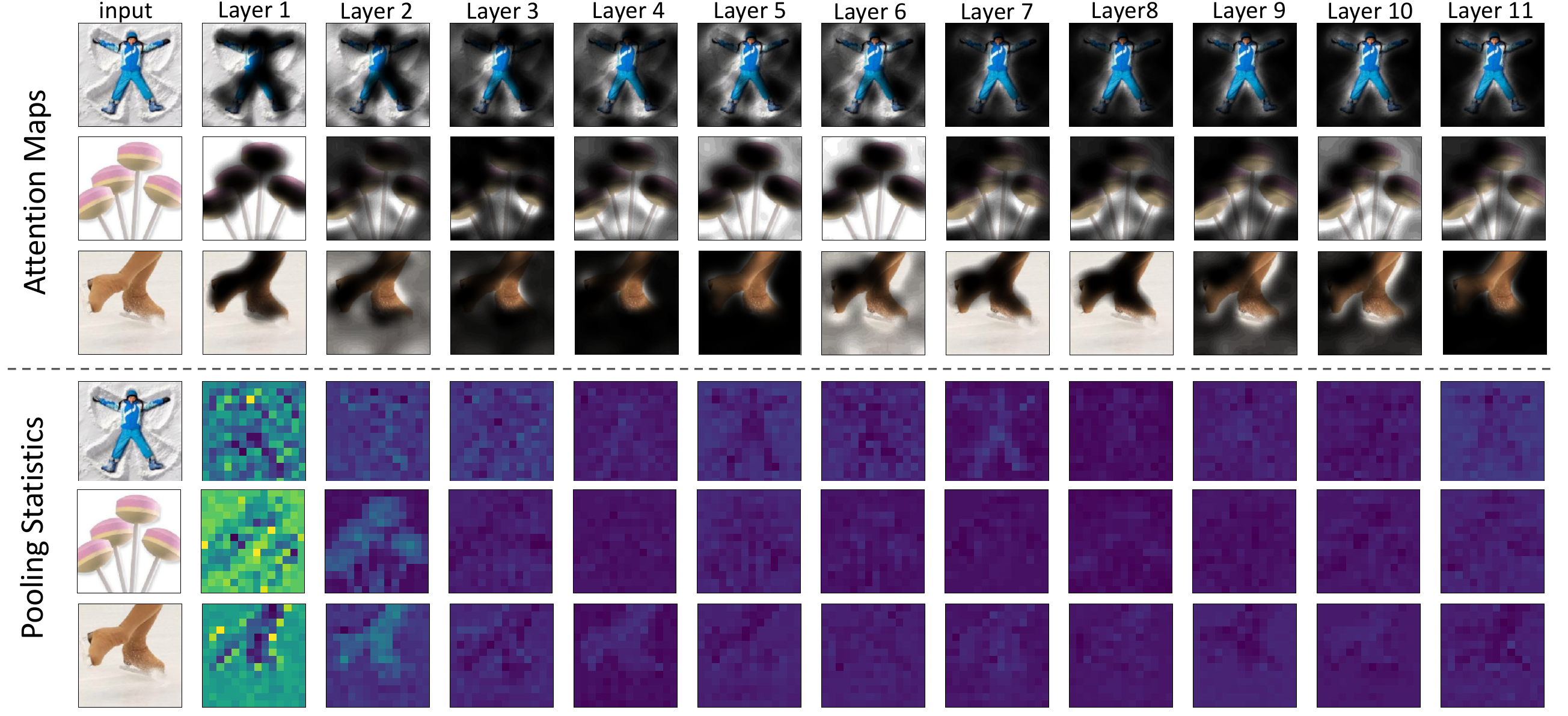}
    \vspace{-23pt}
    \caption{Contribution of different layers in Omnidirectional representations for a given set of examples. On top, we plot the omnidirectional attention maps (using OmniNet$_{B/16}$-P$12$ ) of one of the heads, over all layers, when CLS token in the last layer is used as query. On the bottom, we show the contribution of each layer to the pooling operation of the Omnidirectional module.}
    \label{fig:layer_contributions_viz}
    \vspace{-15pt}
\end{figure*}

We train OmniNet$_{B/32}$ and OmniNet$_{B/16}$ with different partition sizes: $P=\{1, 2, 4, 6, 12\}$. Partition size $P=1$ is simply having no vertical attention and it is just replacing normal attention in ViT, with Performer. We compare these models in terms of their linear 5-shot accuracy on ImageNet dataset (similar to the ablation studies in~\citep{dosovitskiy2020image}). Figure~\ref{fig:OmniNet-ViT-ablation} presents the performance of each model as well as their computational cost during pre-training.

A few patterns can be observed. For both OmniNet$_{B/32}$ and OmniNet$_{B/16}$, the power of omnidirectional directional representations kicks in when we work with partition sizes of more than $2$.  The input resolution during pre-training is $224\times224$, so for $/32$ and $/16$ models the input sequence length to the model is $49$ and $196$. So when setting $P=1$ or $P=2$,  with such sequence lengths, when using an efficient attention engine, like Performer, which provides an approximation of the dot-product-attention, we do not gain a lot on the speed and we lose a bit of performance. However, when using a larger partition size, the additional compute with respect to the performance gain becomes reasonable. 

In both $/32$ and $/16$, the computation cost is almost similar for $P=4$ and $P=6$. With $P=4$, we have three omnidirectional attention, each applied on $4$ layers, while with $P=6$ we have two omnidirectional attention, each applied on $6$ layers. However, $P=6$ gives slightly better results in terms of accuracy and is placed on a sweeter spot in this trade-off. 
With $P=12$, the computational costs of OmniNet increase, but the gain in the performance helps the model to be on the frontier of the compute-performance trade-off, when it is compared to  OmniNet$_{B/32}$ and OmniNet$_{B/16}$.

\subsection{Visualization}

OmniNet combines information from different layers via two sequential mechanisms (\S\ref{sec:omni_rep}): (1) omnidirectional attention, where representations of all tokens in all layers get updated with respect to each other using an efficient attention mechanism; and (2) a pooling operation, where for each token, we collect the best values from all layers. 

In order to understand how these two mechanisms combine information across different layers, we visualize attention maps~\citep{abnar-zuidema-2020-quantifying} and pooling statistics for a set of examples in the image recognition task. 
Figure~\ref{fig:layer_contributions_viz} depicts three example inputs, where we show how OmniNet attends to different layers, as well as each layer's contribution during the pooling operation.

We can see that in some layers, attention seems to detect the objects in the image via attending to the edges or specific parts of the object, while in other layers, the attention mechanism uses mostly background information. It is clear that omnidirectional attention does indeed use such information by actively attending to layers of varying depth.

Additionally, when performing the element-wise pool operation over all the layers for each token, only a fraction of values from each layer's representation make it to the final representation. The bottom rows in Figure~\ref{fig:layer_contributions_viz} illustrate this fraction for each token (image patch) across different layers. In most examples, we observe that a majority of the representation after the pooling operation comes from the first few layers. This is further evidence of how OmniNet can provide an explicit path for directing fine-grained information that is captured by the early layers to the final output, leading to much richer representations.
For the sake of brevity, we refer readers to the Appendix for more detailed plots for these examples as well as other examples, which illustrate the same trends.
\vspace{-6pt}
\section{Conclusion}
\vspace{-1pt}
In this paper, we proposed OmniNet, which uses omnidirectional attention to connect all tokens across the entire network via self-attention. In order to manage the computational costs of the full receptive field, the meta-learner in OmniNet is parameterized by fast and efficient self-attention models. The proposed method achieves stellar performance on a myriad of language and vision tasks. Concretely, OmniNet achieves state-of-the-art performance on WMT EnDe and EnFr, outperforming deep 60-layer transformers. OmniNet also demonstrates substantial improvement over ViT on image recognition tasks. 

\bibliography{ref}
\bibliographystyle{icml2020}

\newpage
\appendix
\section{Detailed Experimental Setup}
This section describes several details of our experiments.
\subsection{Dataset Specific Setups}
For all experiments, we implement code using Python and JAX. Specifically, the main Transformer blocks and codebase for most experiments are derived from FLAX examples. For WMT'17, we build sentencepiece tokenizers of $32K$ from the dataset. WMT'17 collections are obtained from Tensorflow datasets (TFDS). For autoregressive language modeling, the C4 corpus is similarly found in TFDS. For both LM1B and C4 tasks, we use a sentencepiece vocab of $32K$ subwords.

\subsection{Efficient Transformer Hyperparameters} For Xformers (efficient transformers), we use implementations derived from FLAX\footnote{\url{https://github.com/google/flax}.}.For Linformer, we use $k=32$ for the low-rank down projection with shared parameters for both key and value. For Performer, we use the default setup from the official implementation. This corresponds to the generalized attention with ReLU activations. We do not use any random features. For BigBird, our codebase similarly links to the official implementation and use the default hyperparameters. The block size is $64$ for BigBird and the number of random blocks is $3$.

\section{Visualisation of Contributions of Layers in Omnidirectional Representations}
\label{app:vis}
\begin{figure*}
\vspace{-10pt}
\centering
\includegraphics[width=0.9\linewidth]{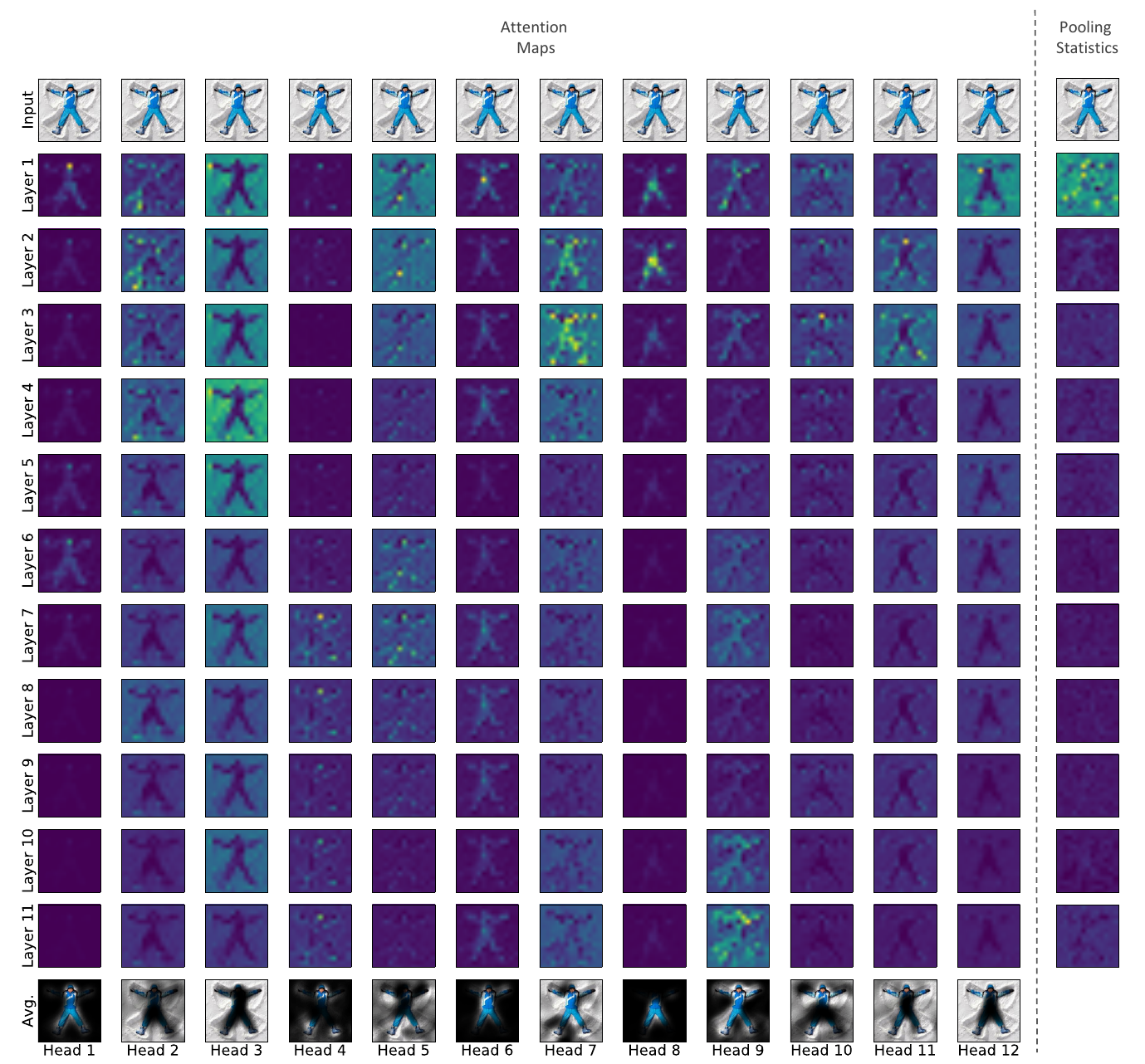}
\vspace{-10pt}
\caption{Contributions of different layers in omnidirectional representations for Example \#1.}
\label{fig:attn_maps_1}
\end{figure*}

\begin{figure*}
\centering
\includegraphics[width=0.9\linewidth]{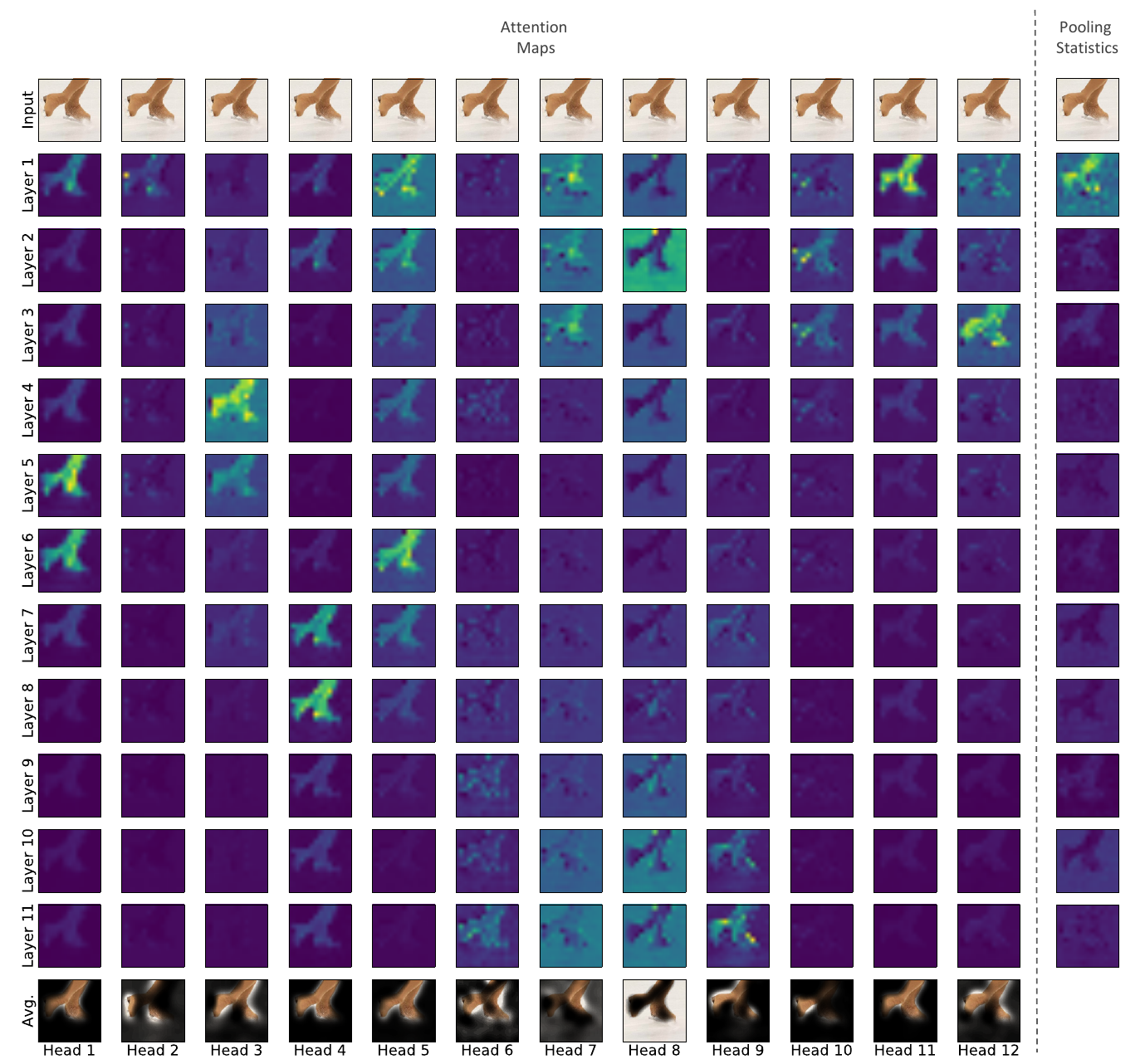}
\vspace{-10pt}
\caption{Contributions of different layers in omnidirectional representations for Example \#2.}
\label{fig:attn_maps_2}
\end{figure*}

\begin{figure*}
\vspace{-10pt}
\centering
\includegraphics[width=0.9\linewidth]{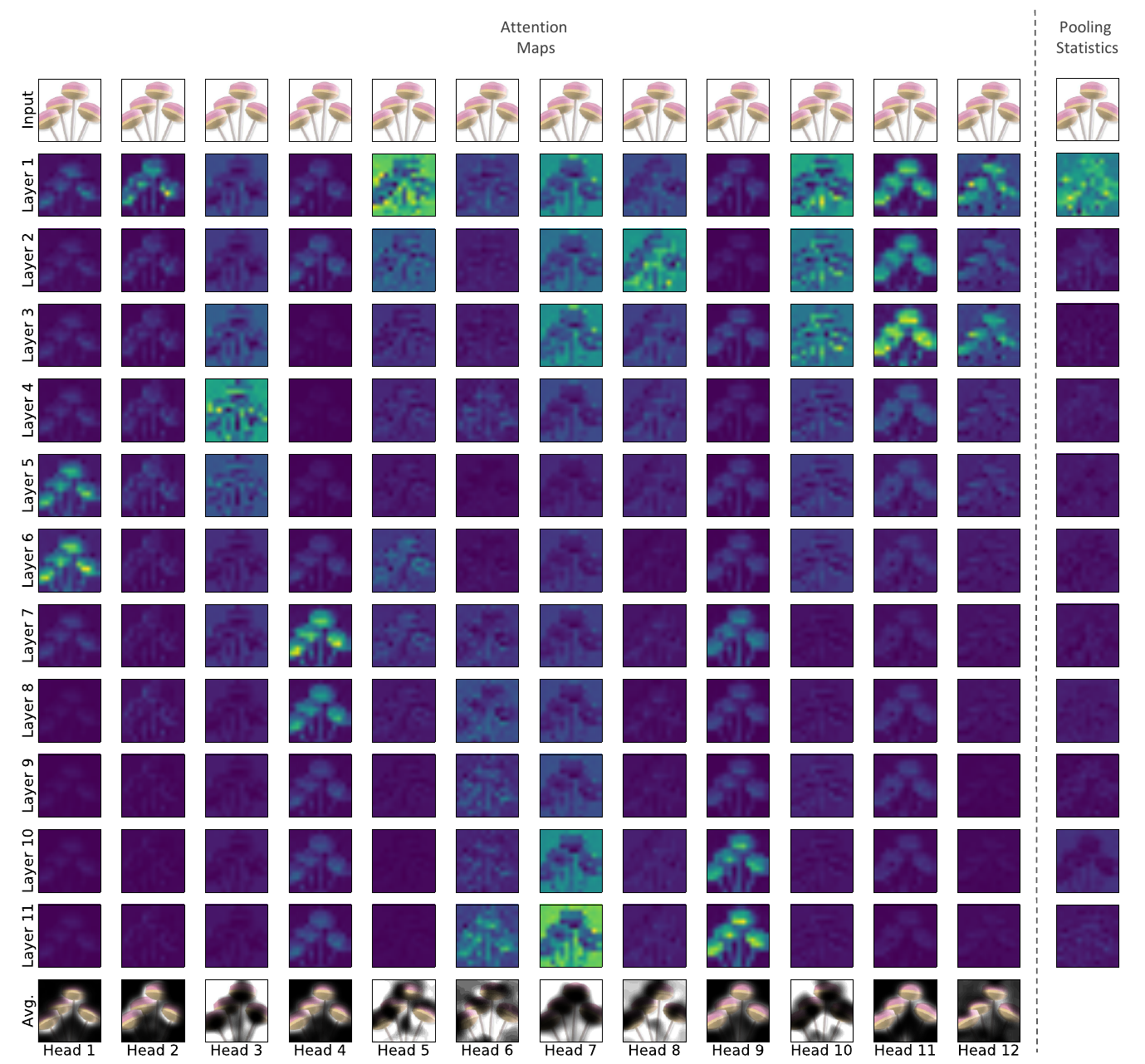}
\vspace{-10pt}
\caption{Contributions of different layers in omnidirectional representations for Example \#3.}
\label{fig:attn_maps_3}
\end{figure*}

\begin{figure*}
\centering
\includegraphics[width=0.9\linewidth]{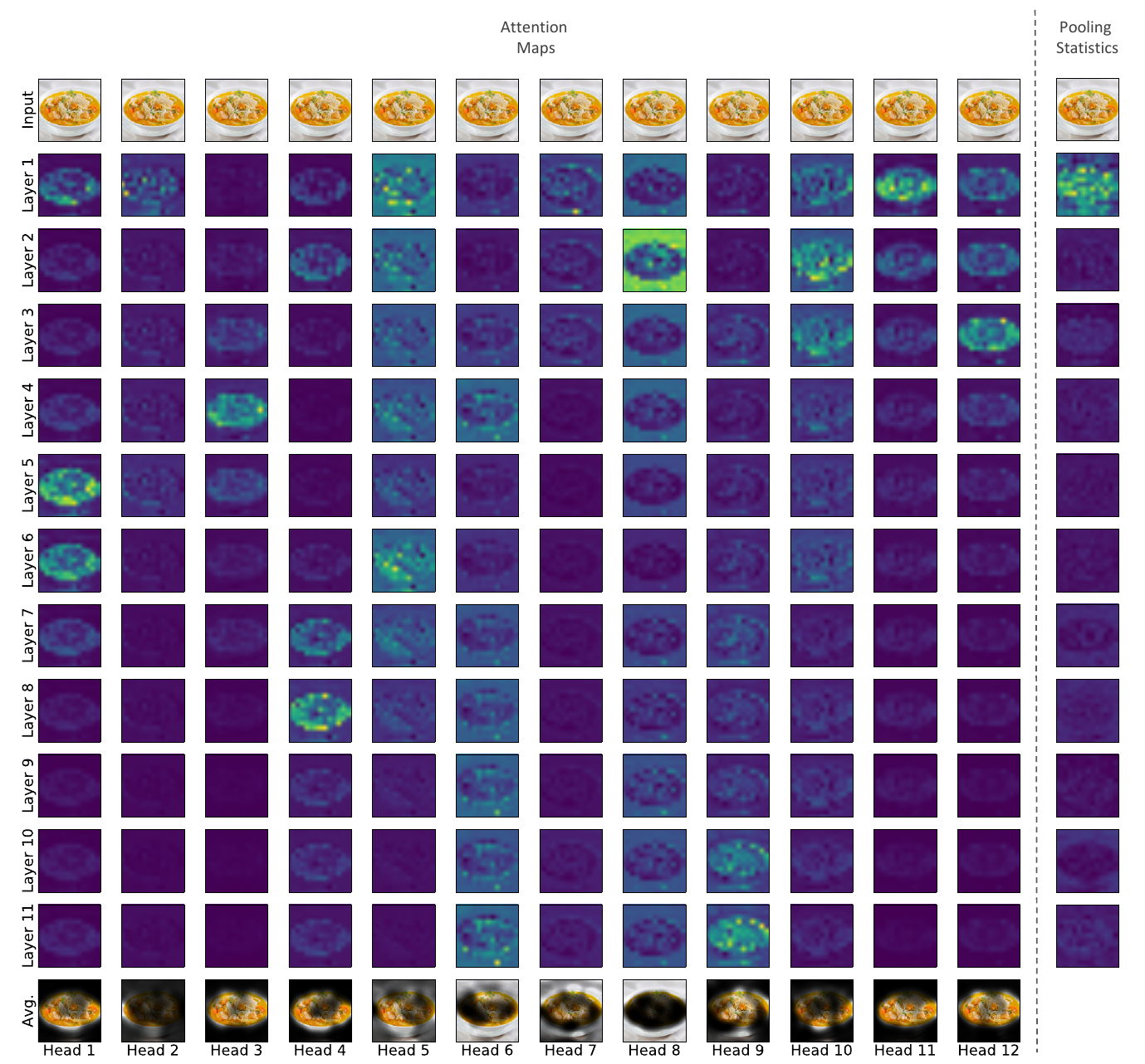}
\vspace{-10pt}
\caption{Contributions of different layers in omnidirectional representations for Example \#4.}
\label{fig:attn_maps_4}
\end{figure*}

\begin{figure*}
\vspace{-10pt}
\centering
\includegraphics[width=0.9\linewidth]{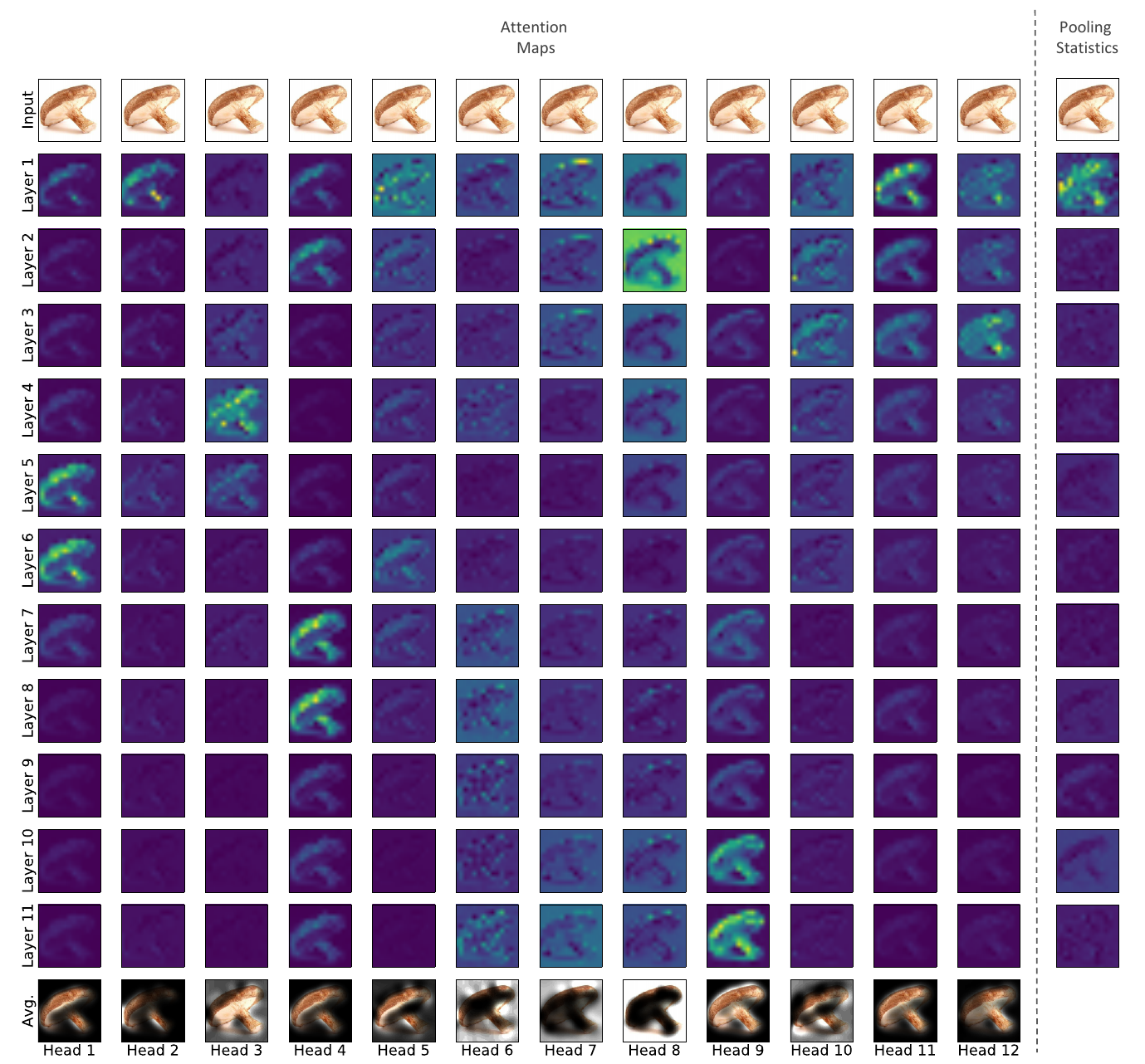}
\vspace{-10pt}
\caption{Contributions of different layers in omnidirectional representations for Example \#5.}
\label{fig:attn_maps_5}
\end{figure*}

Figures~\ref{fig:attn_maps_1} to ~\ref{fig:attn_maps_5} (in subsequent pages) show contributions of different layers in omnidirectional representations in terms of detailed attention maps (attention distribution over all layers, in all heads, when the CLS token in the omnidirectional layer is considered as the query) as well as contribution of different layers in the pooling operation.


\end{document}


\maketitle

\appendix
\section{Omnidirectional Attention Visualizations}
Figures \ref{fig:candy_attn_map} - \ref{fig:two_bottles_attn_map} show detailed omnidirectional attention maps.

\begin{figure*}
    \centering
    \includegraphics[width=\linewidth]{figs/attn_maps/candy.png}
    \caption{}
    \label{fig:candy_attn_map}
\end{figure*}

\begin{figure*}
    \centering
    \includegraphics[width=\linewidth]{figs/attn_maps/car.png}
    \caption{}
    \label{fig:car_attn_map}
\end{figure*}

\begin{figure*}
    \centering
    \includegraphics[width=\linewidth]{figs/attn_maps/dog.png}
    \caption{}
    \label{fig:dog_attn_map}
\end{figure*}

\begin{figure*}
    \centering
    \includegraphics[width=\linewidth]{figs/attn_maps/dog_playing.png}
    \caption{}
    \label{fig:dog_playing_attn_map}
\end{figure*}

\begin{figure*}
    \centering
    \includegraphics[width=\linewidth]{figs/attn_maps/eye.png}
    \caption{}
    \label{fig:eye_attn_map}
\end{figure*}

\begin{figure*}
    \centering
    \includegraphics[width=\linewidth]{figs/attn_maps/fish.png}
    \caption{}
    \label{fig:fish_attn_map}
\end{figure*}

\begin{figure*}
    \centering
    \includegraphics[width=\linewidth]{figs/attn_maps/ice_skating.png}
    \caption{}
    \label{fig:ice_skating_attn_map}
\end{figure*}

\begin{figure*}
    \centering
    \includegraphics[width=\linewidth]{figs/attn_maps/man_pointing.png}
    \caption{}
    \label{fig:man_pointing_attn_map}
\end{figure*}

\begin{figure*}
    \centering
    \includegraphics[width=\linewidth]{figs/attn_maps/three_women_striped_shirt.png}
    \caption{}
    \label{fig:three_women_attn_map}
\end{figure*}

\begin{figure*}
    \centering
    \includegraphics[width=\linewidth]{figs/attn_maps/two_bottles.png}
    \caption{}
    \label{fig:two_bottles_attn_map}
\end{figure*}